\documentclass[review]{elsarticle}

\usepackage{lineno,hyperref}
\modulolinenumbers[5]

\journal{Pattern Recognition}

\usepackage[figuresright]{rotating}
\usepackage{amsopn}
\usepackage{amssymb}
\usepackage{amsmath}
\usepackage{subfig}
\usepackage{array}
\usepackage{tabularx}
\usepackage{multirow}
\usepackage{booktabs}
\usepackage{graphicx}
\usepackage{mathtools}
\usepackage{algorithmic}
\usepackage{algorithm}
\usepackage{listings}
\usepackage{setspace}
\usepackage{url}
\usepackage{float}
\usepackage{epsf} 
\usepackage{tikz}
\usepackage{amsthm}
\usepackage{color}
\usepackage{lineno}
\usepackage{todonotes}
\usepackage[export]{adjustbox}
\theoremstyle{definition}
\usepackage{url}
\usepackage{mathtools}
\DeclarePairedDelimiterX{\norm}[1]{\lVert{\,}}{{\,}\rVert}{#1}

\newcommand{\etal}{\textit{et al.}}
\newcommand{\ie}{\textit{i.e.}}

\newcommand{\viz}{\textit{viz.}}

\newcommand{\myparagraph}[1]{\vspace{0pt}\noindent{\bf #1}}

\newtheorem{definition}{Definition}[section]

\makeatletter
\newcommand{\pushright}[1]{\ifmeasuring@#1\else\omit\hfill$\displaystyle#1$\fi\ignorespaces}
\newcommand{\pushleft}[1]{\ifmeasuring@#1\else\omit$\displaystyle#1$\hfill\fi\ignorespaces}
\makeatother

\DeclareGraphicsExtensions{.pdf,.png}
\graphicspath{{./figures/}}

\begin{document}

\begin{frontmatter}

\title{Learning Graph Edit Distance by Graph Neural Networks}

\author[cvc]{Pau Riba\corref{cor}}
\ead{priba@cvc.uab.cat}
\author[unifr,uasaws]{Andreas Fischer}
\ead{andreas.fischer@unifr.ch}
\author[cvc]{Josep Llad\'{o}s}
\ead{josep@cvc.uab.cat}
\author[cvc]{Alicia Forn\'{e}s}
\ead{afornes@cvc.uab.cat}
\cortext[cor]{Corresponding author}
\address[cvc]{Computer Vision Center, Universitat Aut\`{o}noma de Barcelona, Spain}
\address[unifr]{Department of Informatics, DIVA Group, University of Fribourg, Fribourg, Switzerland}
\address[uasaws]{Institute of Complex Systems, University of Applied Sciences and Arts Western Switzerland, Fribourg, Switzerland}

\begin{abstract}
    The emergence of geometric deep learning as a novel framework to deal with graph-based representations has faded away traditional approaches in favor of completely new methodologies. In this paper, we propose a new framework able to combine the advances on deep metric learning with traditional approximations of the graph edit distance. Hence, we propose an efficient graph distance based on the novel field of geometric deep learning. Our method employs a message passing neural network to capture the graph structure, and thus, leveraging this information for its use on a distance computation. The performance of the proposed graph distance is validated on two different scenarios. On the one hand, in a graph retrieval of handwritten words~\ie~keyword spotting, showing its superior performance when compared with (approximate) graph edit distance benchmarks. On the other hand, demonstrating competitive results for graph similarity learning when compared with the current state-of-the-art on a recent benchmark dataset.
\end{abstract}


\end{frontmatter}


\section{Introduction}
\label{s:intro}

Graph-based representation have been widely used in several application domains such as computer vision~\cite{yang2018rcnn}, bioinformatics~\cite{gilmer2017neural} or computer graphics~\cite{gkioxari2019iccv}. Graphs are powerful and flexible representations able to describe shapes, images, knowledge, etc. in terms of relationships between constituent parts or primitives. In the core of any pattern recognition application, there is the ability to compare two objects. This operation, which is trivial when considering feature vectors defined in $\mathbb{R}^n$, is not properly defined in the graph domain~\cite{Conte2004,Foggia2014}. Due to the inherent graph flexibility, it forces us to adopt some definitions of dissimilarity (similarity) ad hoc to particular purposes. Borgwardt~\cite{borgwardt2007thesis} formally defines such problem as follows:

\begin{definition}[Graph Comparison Problem]
    Let \(g_1=(V_1,E_1,\mu_1,\nu_1)\) and \(g_2=(V_2,E_2,\mu_2,\nu_2)\) be two graphs from \(\mathcal{G}\), the graph comparison problem is to find a function
    \[
        d \colon \mathcal{G} \times \mathcal{G} \to \mathbb{R}
    \]
    such that \(d(g_1,g_2)\) quantifies the dissimilarity (similarity) of \(g_1\) and \(g_2\).
\end{definition}

Lots of efforts have been made in this direction. In the literature, error-tolerant or inexact graph matching algorithms have been proposed. For instance, Zhou and de la Torre~\cite{zhou2015fgz} proposed to factorize the large pairwise affinity matrix into smaller matrices that encode, on the one hand, the local node structure of each graph, and on the other hand, the pairwise affinity between both nodes and edges. Moreover, graph embeddings and kernels have been also proposed as mechanisms to compare two graphs. Graph embedding refers to those techniques that aim to explicitly map graphs to vector spaces~\cite{Gibert2012,dutta2018hierarchical}. Similarly, implicit graph embedding or graph kernel aims at finding a function able to map the input graph into a \emph{Hilbert space} which basically defines a way to compute the similarity between two graphs in terms of a dot product~\cite{borgwardt2007thesis, Kondor2016}. One of the most popular error-tolerant graph matching methods is the graph edit distance (GED)~\cite{sanfeliu1983distance,bunke1983inexact,gao2010survey}. Now, the graph comparison problem is formulated in terms of finding the minimum transformation cost in such a way that an isomorphism exists between the transformed graph $g_1$ and the second one $g_2$. In addition, GED algorithms, unlike most embedding and kernel methods, are able to cope with any type of labeled graph and any type of labels on nodes and edges.

The main drawback of GED techniques is that the time complexity is exponential in terms of the number of nodes of the input graphs. Hence, GED is unfeasible in a real scenario, where there may be no constraints in terms of the graph size. Therefore, several algorithms have been proposed to cope with this complexity~\cite{riesen2009approximate,fischer2015hausdorff}. However, these approximate algorithms only consider very local node structures in their computation and they do not adapt their costs according to the problem being addressed.

In Euclidean domains such as vectors, images or sequences, deep learning has been proposed as a solution to perform a huge variety of tasks. In the last decade, the particular case of deep neural networks have supposed a breakthrough in computer vision, pattern recognition and artificial intelligence~\cite{krizhevsky2012imagenet}. However, until recent years, deep learning advances were not able to process non-Euclidean data in its framework. Lately, \emph{geometric deep learning}\footnote{\url{http://geometricdeeplearning.com/}} has emerged as a generalization of deep learning methods to non-Euclidean domains such as manifolds and graphs~\cite{bronstein2017geometric}. This field has arisen in the recent years allowing the developed models to encode structural and relational data. Several fields have benefited from this new paradigm, for instance computer vision~\cite{yang2018rcnn}, quantum chemistry~\cite{gilmer2017neural} and computer graphics~\cite{gkioxari2019iccv} among others.

Inspired by the efficient GED approximations and the powerful framework provided by the new advances in geometric deep learning, we propose to leverage its effectiveness as a learning framework to enhance a graph distance computation. Therefore, we are facing a graph metric learning problem. 
It can be formulated as a contrastive learning problem that finds contrast between similar and dissimilar objects. A siamese architecture is suitable for this problem. Bromley~\etal~\cite{bromley1994signature} proposed it for signature verification. 
Siamese networks make use of the same model and weights on two separated branches in order to learn a representation where distances can be computed. Later, several approaches have extended this idea, being the triplet loss~\cite{weinberger2009distance} one of the most successful methods. Recently, novel approaches have focused on extending this concept in order to exploit groups of samples instead of pairs or triplets~\cite{elezi2019group}. Moreover, contrastive learning has been used, not only as a metric learning framework but it has also raised some attention due to its astonishing improvement on unsupervised learning tasks~\cite{chen2020simple}.

In this work, we propose to define a triplet learning framework for the graph metric learning problem. In our proposed approach an enriched graph representation is learned by means of a graph neural network. In addition, our proposed distance is based on the Hausdorff Edit Distance introduced by Fischer~\etal~\cite{fischer2015hausdorff} as an efficient approximation of the real graph edit distance. In comparison, our framework has the ability to enrich the initial graph representation by means of message passing operations which learns the edit cost operations. Furthermore, insertion and deletion costs, are dynamically learned according to nodes local contexts. Therefore, we avoid a costly manual process on setting the edit cost operations per each specific problem.

The proposed approach is validated using standard graph datasets for keyword spotting and object classification. In this application scenario, the proposed approach based on a message passing neural network shows competitive results demonstrating the efficacy of our learning framework.

This article supposes a significantly extended version of our previous conference paper~\cite{riba2018learning}. To the best of our knowledge, it was the first work that introduced the idea of learning a graph metric by means of message passing architectures. In this work, we have enhanced our previous graph neural network architecture. Specifically, the main changes from our previous work are (i) a new graph neural network architecture to obtain a better node context representation; (ii) a novel graph similarity which makes use of learned insertions, deletions and substitutions as edit operations; (iii) a learning approach making use of triplets of graphs with in-triplet hard negative mining. Finally, a thorough analysis and evaluation of the involved parameters as well as a performance comparison with the recent state-of-the-art literature is presented.

The rest of this paper is organized as follows. Section~\ref{s:related} introduces the related work on graph neural networks. Section~\ref{s:ged} introduce the graph edit distance algorithm along with relevant approximations. Section~\ref{s:method} and~\ref{s:learn} proposes our learning framework and learning strategy. Section~\ref{s:validation} evaluates the proposed framework. Finally, Section~\ref{s:conc} draws the conclusions and future work.

\section{Related Work on Geometric Deep Learning}
\label{s:related}

In the following, some basic notations, definitions and previous works are presented in the context of geometric deep learning. As mentioned in the introduction, \emph{Geometric deep learning} has emerged as a generalization of deep learning methods to non-Euclidean domains such as graphs and manifolds~\cite{bronstein2017geometric}. In this document we will specifically focus on its applications to graphs.

A graph can be roughly defined as a symbolic data structure describing relations (\emph{edges}) between a finite set of objects (\emph{nodes}). Graphs are formally defined as

\begin{definition}[Graph]
    \label{def:graph}
    Let \(L_V\) and \(L_E\) be a finite or infinite label sets for nodes and edges, respectively. A \emph{graph} \(g\) is a \(4-\text{tuple}\) \(g=(V,E,\mu,\nu)\) where,
    \begin{itemize}
        \item \(V\) is the finite set of \emph{nodes}, also known as \emph{vertices},
        \item \(E\subseteq V\times V\) is the set of \emph{edges},
        \item \(\mu \colon V \to L_V\) is the node labelling function, and
        \item \(\nu \colon E \to L_E\) is the edge labelling function.
    \end{itemize}
\end{definition}

We denote \(|V|\) and \(|E|\) as the \emph{order} and \emph{size} of a graph, namely, the number of nodes and edges respectively. Moreover, the \emph{neighborhood} of a given node \(v \in V\) in a graph \(g\) is defined as the set of nodes $\{v_i\}$ adjacent to \(v\). We denote the neighborhood as \(\mathcal{N}(v)\).

\subsection{Graph neural networks}

Graph Neural Networks were first proposed by Gori and Scarselli~\cite{gori2005graph,scarselli2009graph} as the first attempts to generalize neural networks to graphs. Later, Bruna~\etal~\cite{bruna2013spectral} proposed the first formulation of CNNs on graphs taking advantage of the graph spectral domain. However, the ability to process graph data came with a huge time complexity becoming inappropriate for real scenarios. Later, the works of Henaff~\etal~\cite{henaff2015deep}, Defferrard~\etal~\cite{defferrard2016convolutional} and Kipf~\etal~\cite{kipf2017semi} addressed these computational drawbacks. In its simplest form, a GNN layer is defined as

\begin{equation}
  h^{(k+1)} = G_c(h^{(k)}) = \rho \left( \sum_{B\in \mathcal{A}^{(k)}}Bh^{(k)} \Theta_B^{(k)} \right),  
  \label{eq:gnn:simple}
\end{equation}
where \(h^{(k)}\) is the node hidden state at the $k$-th layer, \(\rho\) is a non-liniarity such as \(\operatorname{ReLU}(\cdot)\), \(\mathcal{A}\) is a set of graph intrinsic linear operators that act locally on the graph signal and \(\Theta\) are learnable parameters. The set of graph intrinsic linear operators can handle multi-relational graphs, however, in most cases, \(\mathcal{A}\) only contains the adjacency matrix.

Recently, Gilmer~\etal~\cite{gilmer2017neural} proposed an approach named \emph{Message Passing Neural Networks} (MPNNs) as a general supervised learning framework for graphs. This approach is able to generalize several GNN layers to a common pipeline. They propose to define each layer by means of two differentiable functions, on the one hand, a message function \(M^{(k)}(\cdot)\) which collects the information from the neighboring nodes and edges according to
\begin{equation}
    m_v^{(k+1)} = \sum_{u\in\mathcal{N}(v)} M^{(k+1)}(h_v^{(k)}, h_u^{(k)}, e_{vu}),
\end{equation}
where \(h_u^{(k)}\) and \(h_v^{(k)}\) are the hidden states of nodes \(v\) and \(u\) at iteration \(k\) and \(\mathcal{N}(v)\) denotes the neighbours of \(v\) in the graph \(g\). On the other hand, an update function which updates the hidden state of the central node \(v\) according to the message \(m_v^{(k+1)}\). The update function is formally defined as
\begin{equation}
    h_v^{(k+1)} = U^{(k+1)}(h_v^{(k)}, m_v^{(k+1)}).
\end{equation}

Several important GNN layers have been proposed along the last years, the most relevants for the scope of this work are the Graph Attention Networks (GAT)~\cite{velivckovic2017graph} and the Gated Graph Neural Networks (GG-NN)~\cite{li2016gated}.

The literature on graph neural networks and their applications is 
quite large, 
 so we refer the interested readers to recent survey papers for a comprehensive overview of these methodologies~\cite{bronstein2017geometric,zhou2018graph,battaglia2018relational,wu2020comprehensive}. Very recently, Dwivedi~\etal~\cite{dwivedi2020benchmarking} presented a reproducible benchmarking framework.

\subsection{Graph metric learning}

Neural networks have been widely used as the learning framework for similarity problems. Promptly, siamese neural networks were adopted as a family of neural networks consisting of two networks with shared weights for similarity learning. For example, Baldi~\etal~\cite{baldi1993neural} makes use of a siamese model for fingerprint recognition whereas Bromley~\etal~\cite{bromley1994signature} presented a siamese architecture for signature verification. Siamese neural networks use a pair of samples to train with positive and negative examples \ie~being similar or not. Later, several approaches have extended this idea in order to take always into account positives and a negatives examples. These approaches are known as 
triplet networks~\cite{weinberger2009distance}. In this case, three networks with shared weights are used to bring similar examples together and dissimilar examples to be far apart.

Inspired on these works, several papers appeared extending these ideas to the graph domain. Thus, we proceed to review a handful of approaches facing the graph similarity learning problem. Li~\etal~\cite{li2019graph} presented two different models to solve the graph similarity problem, on the one hand a graph embedding model, and on the other hand, a graph matching network. Both models can be trained with pairs or triplets. Let us briefly review each one of these models,

\begin{itemize}
    \item \textbf{Graph embedding model}: This model, takes advantage of siamese GNN's to embed the given graphs into a vectorial space. Then, given the pair of vectorial representations, a similarity metric in the vector space can be computed by means of the Euclidean, cosine or Hamming similarities. Very similar approaches have also been presented in other works, for instance Chaudhuri~\etal~\cite{chaudhuri2019siamese} who trained a similar approach with contrastive loss or the work introduced by Zhang~\etal~\cite{zhang2020graph} which uses the \(L_2\) loss to mimic the real similarity score.
    
    \item \textbf{Graph matching networks (GMN)}: Similarly to the previous architecture, two GNNs with shared weights process the input graphs. However, in this case, the authors propose to modify the node update module in order to take into account not only the aggregated messages on the edges of each graph, but a cross-graph message which measures how well the nodes match from one graph to the other. Finally, following the same idea as the previous model, each graph is finally converted into a vectorial representation which is later used in a similarity metric.
\end{itemize}

Another interesting approach, namely SimGNN, was proposed by Bai~\etal~\cite{bai2019simgnn}. In this work, the authors proposed to combine graph-level embeddings and node-node similarity scores by taking their histogram of features. However, as the histogram function is not differentiable, this methodology still relies on the graph-level embedding for computing the final similarity score. Their model is trained according to the real graph edit distance for small datasets whereas the smallest distance computed by three well-known approximate algorithms is taken to handle large datasets.  The authors extended this work by proposing a new model named GraphSim~\cite{bai2020learning}. In this architecture, only three node-node similarities scores are used corresponding to node embeddings at different scales. After that, the similarity matrices are treated as images and a CNN is used to process them to discover the optimal node matching pattern. However, to deal with the permutation invariant ordering of graph nodes, they propose a BFS ordering. It also allows the use of CNN as they claim that the required spatial locality performs properly. Moreover, the similarity matrices are first padded to \(\max(|V_1|,|V_2|)\), where \(V_1\) and \(V_2\) are the node sets of the graphs involved and resized to meet the expected size. Finally, following the SimGNN training, a precomputed similarity score is used to lead the training. 

Compared to these works, our model makes use of a node-node distance matrix to obtain a global graph distance metric. Therefore, we are not obtaining a global vectorial representation of our graphs nor applying cross-convolution layers in our graph neural network architecture. This allows us to make use of any differentiable graph or set distance, which preserves the permutation invariance property, while avoiding the computational overhead of the cross-convolution layers. Moreover, we avoid the loss of structural information of other approaches when obtaining a vectorial graph representation by explicitly dealing with the structure in the distance itself.

\section{Related work on Graph Edit Distance}
\label{s:ged}

This Section introduces the concept of graph edit distance (GED) jointly with relevant cubic and quadratic time approximations.

\subsection{Definition}

The \emph{Graph Edit Distance} (GED)~\cite{gao2010survey,Sanfeliu1983,bunke1983inexact} evaluates the similarity of two graphs in terms of edit operations. The GED is inspired by the \emph{String Edit Distance} (SED), also known as the \emph{Levenshtein} distance~\cite{levenshtein1966binary,wagner1974string}. Actually, it can be seen as a generalisation of SED as strings can be viewed as a special case of graphs. In this specific case, the order of the characters allows the efficient use of dynamic programming to find the string-to-string correspondence. However, for general graphs, the correspondence cannot rely on a specific ordering of nodes and edges.

The main idea of GED is to compute the minimum cost transformation from the source graph \(g_1\) to the target one \(g_2\) in terms of a sequence of edit operations \(e_1,\ldots,e_k\). This sequence of edit operations is named \emph{edit path} between \(g_1\) and \(g_2\). GED is formally defined as

\begin{definition}[Graph Edit Distance]
    Let \(g_1=(V_1,E_1,\mu_1,\nu_1)\) and \(g_2=(V_2,E_2,\mu_2,\nu_2)\) be the source and the target graphs respectively. The \emph{graph edit distance} between \(g_1\) and \(g_2\) is defined by
    \[
    d(g_1,g_2) = \min_{(e_1,\ldots,e_2) \in \Upsilon(g_1, g_2)} \sum_{i=1}^k c(e_i),
    \]
    where \(\Upsilon(g_1,g_2)\) denotes the set of edit paths transforming \(g_1\) into \(g_2\), and \(c(e_i)\) denotes the cost function measuring the strength of the edit operation \(e_i\).
\end{definition}

Usually the considered edit operations are \emph{insertion}, \emph{deletion} and \emph{substitutions} for nodes and edges. For instance, a frequently used cost functions for nodes and edges with labels defined in $\mathbb{R}^n$ is

\begin{equation}
\begin{split}
    \mathcal{C}(u\longrightarrow v)&=\alpha \cdot \norm{\mu_1(u) - \mu_2(v)}\\
    \mathcal{C}(u\longrightarrow \varepsilon )&=\alpha \cdot \tau_n ,\quad \mathcal{C}(\varepsilon \longrightarrow v)=\alpha \cdot \tau_n \\
    \mathcal{C}(p\longrightarrow q)&=(1-\alpha) \cdot \norm{\nu_1(p) - \nu_2(q)}\\
    \mathcal{C}(p\longrightarrow \varepsilon )&=(1-\alpha) \cdot \tau_e ,\quad \mathcal{C}(\varepsilon \longrightarrow q)=(1-\alpha) \cdot \tau_e
\end{split}
\end{equation}
for nodes \(u \in V_1\), \(v \in V_2\) and edges \(p \in E_1\), \(q \in E_2\) and user-defined parameters \(0 \leq \alpha \leq 1\) indicating a trade-off between node and edge costs and \(\tau_n, \tau_e > 0\) to fix the deletion and insertion costs for nodes and edges respectively.

The graph edit distance is a known NP-complete problem (see~\cite{zeng2009comparing} for a detailed proof), exponential with respect to the number of nodes. Thus, in addition to exact GED algorithms, some efficient approximations have been proposed~\cite{justice2006binary,neuhaus2006fast}. In the following we will review two techniques that have been widely adopted in the literature.

\subsection{Assignment edit distance}

The \emph{assignment edit distance} (AED), also known as \emph{bipartite graph matching}, proposed by Riesen~\etal~\cite{riesen2009approximate}, is a cubic time approximation of GED with respect to the number of nodes of the involved graphs. It provides an upper bound of order \(\mathcal{O}\left((n_1 + n_2)^3\right)\) where \(n_1 = |V_1|\) and \(n_2 = |V_2|\).

The main idea is to transform the GED computation to an assignment problem between nodes and their local structure. This method, defines a matrix of edit costs between the nodes of both graphs. Afterwards, the best correspondence between nodes is found by a linear assignment method~\cite{munkres1957algorithms}. The matrix definition for the AED algorithm takes into consideration both, the local structure of the vertices and their attributes. The cost matrix \(C\) is defined as
\[
C = \left[
    \begin{array}{c c c | c c c}
        c_{1,1} & \cdots & c_{1,m} & c_{1,\varepsilon} & \cdots & \infty \\
        \vdots  & \ddots & \vdots  & \vdots         & \ddots          & \vdots \\
        c_{n,1} & \cdots & c_{n,m} & \infty         & \cdots          &  c_{n,\varepsilon} \\
        \hline
        c_{\varepsilon,1} & \cdots & \infty          & 0 & \cdots & 0 \\
        \vdots         & \ddots & \vdots          & \vdots & \ddots          & \vdots \\
        \infty         & \cdots           &  c_{\varepsilon,n} & 0 & \cdots          &  0
    \end{array}
\right]
\]
where \(c_{i,j}\) denotes the cost of a node substitution \(c(u_i \rightarrow v_j)\); \(c_{i, \varepsilon}\) denotes the cost of a node deletion \(c(u_i \rightarrow \varepsilon)\); and \(c_{\varepsilon, j}\) denotes the costs of a node insertion \(c(\varepsilon \rightarrow v_j)\) where $v_i \in V_1$ and $v_j \in V_2$. 

Other works have focused on speeding-up this method in some particular settings. For example, Serratosa~\etal~\cite{serratosa2014fast} define an algorithm able to reduce the computation time with the only constrain that the edit costs should define the graph edit distance as a real distance function, that is, the cost of insertion plus deletion of nodes and edges have to be lower or equal than the cost of substitution of nodes and edges.

\subsection{Hausdorff Edit Distance}
\label{ssec:hed}

Despite obtaining a good approximation, the time complexity of the AED algorithm is still a problem for some applications where the time is an important constrain. In order to alleviate this issue, Fischer~\etal~\cite{fischer2015hausdorff} proposed the \emph{Hausdorff Edit Distance} (HED) which is a lower bound approximation of the real GED with a quadratic time complexity of \(\mathcal{O}\left(n_1 \cdot n_2\right)\) where \(n_1 = |V_1|\) and \(n_2 = |V_2|\). HED is based on the \emph{Hausdorff distance} is formally defined as

\begin{definition}[Hausdorff distance]
    Let \(A\) and \(B\) be two non-empty subsets of a metric space \((M,d)\). The Hausdorff distance \(d_H(A,B)\) is defined as
    \[
        d_H(A,B)=\max \left( \sup_{a\in A} \inf_{b\in B} d(a,b), \sup_{b\in B} \inf_{a\in A} d(a,b) \right).
    \]
\end{definition}

For finite sets \(A\),\(B\) the Hausdorff distance is reformulated as
\[
    d_H(A,B)=\max \big( \max_{a\in A} \inf_{b\in B} d(a,b), \max_{b\in B} \inf_{a\in A} d(a,b) \big).
\]

By definition, the Hausdorff distance is very sensitive to outliers. Hence, in their work they propose to replace the maximum operator with the summation operation, which forces the distance to take into account all nearest neighbour distances and becomes more robust to noise than the original one. Thus, the new distance is defined as
\begin{equation}
    \label{eq:classicsoa:hhat}
    d_{\hat{H}}(A,B)=\sum_{a\in A} \min_{b\in B} d(a,b) + \sum_{b\in B} \min_{a\in A} d(a,b).
\end{equation}

Until now, this distance \(d_{\hat{H}}(\cdot)\) do not consider node insertions and deletions. Therefore, from Equation~\ref{eq:classicsoa:hhat} they define a new distance on graphs. Given two graphs \(g_1=(V_1,E_1,\mu_1,\nu_1)\) and \(g_2=(V_2,E_2,\mu_2,\nu_2)\) and a matching cost defined as \(c\), the HED is defined as,
\begin{equation}
    \label{eq:classicsoa:hed}
    \operatorname{HED}(g_1,g_2,c)=\sum_{u\in V_1} \min_{v\in V_2 \cup \{ \varepsilon \}} c_n^*(u,v) + \sum_{v\in V2} \min_{u\in V_1 \cup \{ \varepsilon \}} c_n^*(v,u),
\end{equation}
where \(c_n^*(u,v)\) is a modified node matching cost defined as,
\[
    c_n^* = \begin{cases}
            \frac{c_n(u,v)}{2}, & \text{if } (u\to c) \text{ is a substitution} \\
            c_n(u,v), & \text{otherwise}.
        \end{cases}
\]

This redefinition of the node matching cost is needed because HED does not enforce bidirectional substitutions. Equation~\ref{eq:classicsoa:hed} is composed by a summation, hence, only if both directions are considered, the full cost will be taken into account. The same matching algorithm is considered, if needed, for the edge matching. In this setting, the HED finds an optimal assignment per node instead of a global optimal assignment as pretended by the real GED. 

A typical drawback of GED approximation algorithm is that it only relies on local edge structures rather than global information. Some efforts have been made to improve the performance by increasing the node context at matching time~\cite{fischer2015improving}. However, obtaining a better knowledge on the relation of each node within the graph is still an open issue that we aim to address by the new advances on graph neural networks.

\section{The Learned Graph Distance Framework}
\label{s:method}

This section is devoted to present our proposed learning framework for graph distance. The proposed model learns the Hausdorff edit distance, proposed by Fischer~\etal~\cite{fischer2015hausdorff}. As a learning setting, the proposed architecture can be trained either with pairs or triplets of graphs. Hence, as ground-truth, only the information on whether or not two graphs belong to the same class is required. Note, that in our proposed approach, we do not require the node correspondence information nor the real graph edit distance. Instead, the node assignment is implicitly learned by our system. Moreover, the edit costs for both insertions, deletions and substitutions are learned by our framework. Therefore, we do not require to manually tune these parameters following the traditional pipeline. Although our framework can be trained using pairs, in this work we will focus on the triplet setting. Hence, we make use of three GNN with shared weights.

\begin{figure}
    \centering
    \includegraphics[width=0.7\textwidth]{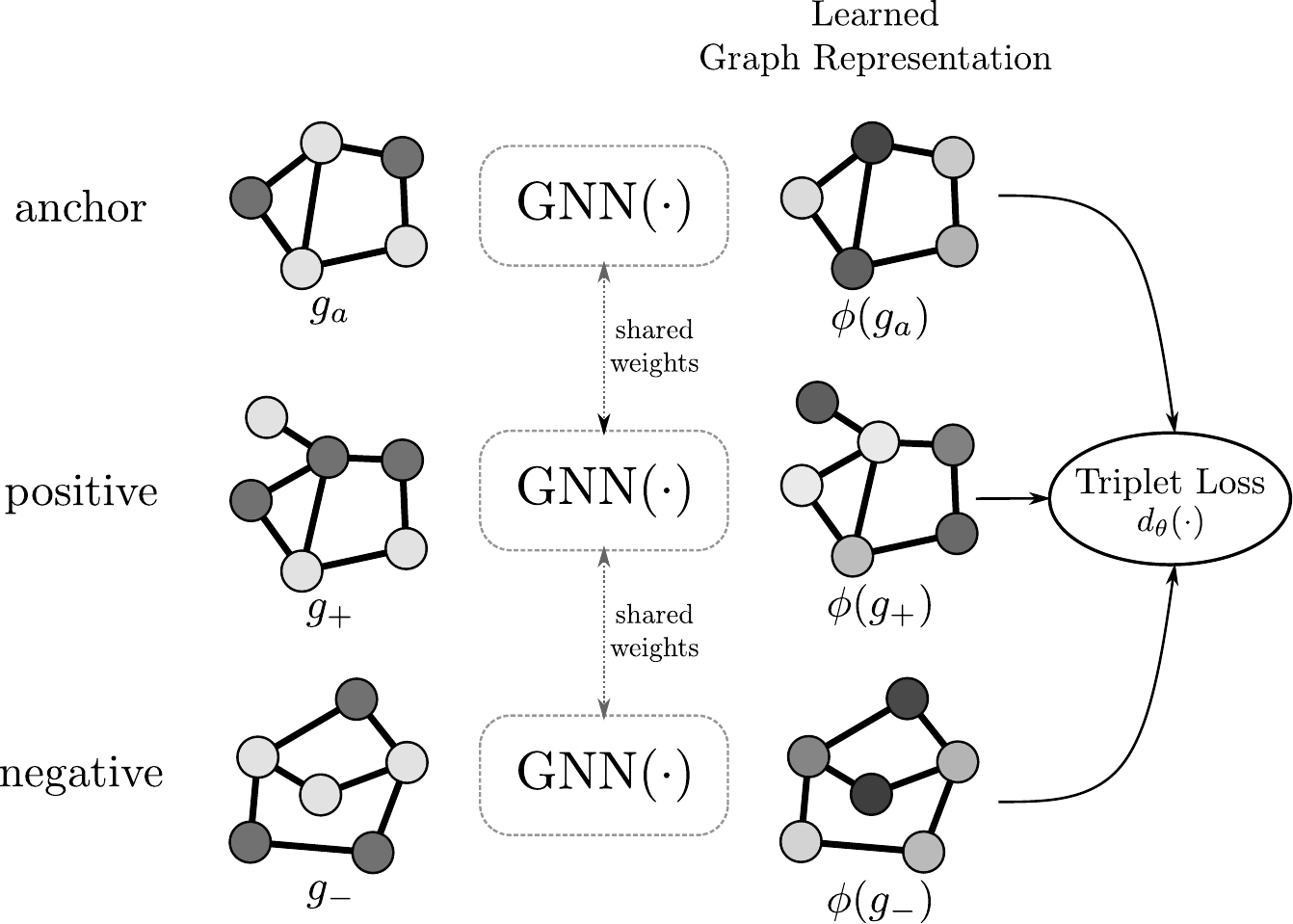}
    \caption{Overview of our learning framework. Given a triplet of graphs \((g_a, g_+, g_-)\) as the anchor, positive and negative samples respectively, the GNN \(\phi(\cdot)\) learns a graph representation per each one \((\phi(g_a), \phi(g_+), \phi(g_-))\), which can be matched by means of a learned distance \(d_{\theta}( \cdot)\).}
    \label{fig:learningdistance:pipeline}
\end{figure}

Figure~\ref{fig:learningdistance:pipeline} shows a graphical outline of our proposed approach. Our pipeline can be divided in two stages. Firstly, a graph neural network \(\phi(\cdot)\) is used to obtain a node-level embedding which codifies the local context information, in terms of structure, for each node. Secondly, a novel graph similarity algorithm based on the Hausdorff edit distance is proposed as a technique to compare two graphs $d_{\theta}(\cdot)$. Observe that the graph similarity can be replaced by any differentiable graph distance approach.

Each stage is carefully described next. First, the GNN architecture is explained in detail. Afterwards, the proposed graph similarity is developed.

\subsection{Learning node embeddings}
\label{ssec:arch}

The first stage of our framework is a graph neural network architecture \(\phi(\cdot)\) able to learn a new graph representation in terms of node embeddings. Our architecture consists of \(K\) propagation layers that map the input graph to an enriched representation. Thus, each propagation layer takes a set of node representations at layer \(k\), \(\{h_i^{(k)}\}_{i \in V}\) and maps it to a new node representation \(\{h_i^{(k+1)}\}_{i \in V}\) at layer \(k+1\).
We evaluate the following two different architectures according to two different message passing strategies.

\myparagraph{GAT-based model}: This model uses graph attention networks (GAT), introduced by Veličković~\etal~\cite{velivckovic2017graph}. A GAT layer is formally defined as
\begin{equation}
    h_v^{(k+1)} = \sum_{u\in \mathcal{N}(v)} \alpha_{vu} W^{(k)} h_u^{(k)},
\end{equation}
where \(\alpha_{vu}\) is the attention score between node \(v\) and node \(u\), \(W^{(k)}\) are the learned weights and \(h_v^{(k)}\) is the hidden state of node \(v\), both, at layer \(k\). The attention score \(\alpha_{vu}\) is learned by
\begin{align}
    \begin{aligned}
        \alpha_{vu}^{(k)} & = \operatorname{softmax_v} (e_{vu}^{l})\\
        e_{vu}^{(k)} & = \operatorname{LeakyReLU}\left(\vec{a}^t [W^{(k)} h^{(k)}_{v} \| W^{(k)} h^{(k)}_{u}]\right),
    \end{aligned}
\end{align}
where $\vec{a}$ and $W^{(k)}$ are a vector and a matrix of learned weights. Moreover, a multi-head attention can be used to enrich the model capacity and to stabilize the learning process.

In our setting, we use these layers with residual connections, four attention heads and BatchNorm layers~\cite{ioffe2015batch} with the exception of the last layer. The attention heads are concatenated at the intermediary layers and averaged for the final layer.

\myparagraph{GRU-based model}: This architecture is based on the gated graph neural networks (GG-NN) proposed by Li~\etal~\cite{li2016gated}. Originally, the message function is formulated as \(M(h_v^{(k)}, h_u^{(k)}, e_{vu}) = A_{e_{vu}}h_u^{(k)}\), where \(A_{e_{vu}}\) is a learned matrix for each possible edge label. Note that we are restricted to a discrete set of labels. In order to overcome this constrain, Gilmer~\etal~\cite{gilmer2017neural} proposed to use a modified message function defined as \(M(h_v^{(k)}, h_u^{(k)}, e_{vu}) = A(e_{vu})h_u^{(k)}\), where \(A(e_{vu})\) is a neural network which maps the edge vector to a matrix \(d\times d\). This modification allows the use of non-discrete information as edge attributes. Finally, the update function is defined as \(U(h_v^{(k)},m_v^{(k)})=\operatorname{GRU}(h_v^{(k)},m_v^{(k)})\), where \(\operatorname{GRU}\) is the Gated Recurrent Unit~\cite{cho2014properties}. In its original formulation, the first node hidden state is padded with zeros to meet the size defined by the \(\operatorname{GRU}\), however, we propose to use a fully-connected layer as a first node embedding. Moreover, we propose to incorporate edge features according to the source and destination node according to, \(e_{vu} = \operatorname{MLP}(|h_v^{(1)} - h_u^{(1)}|)\) as proposed in~\cite{garcia2018fewshot}. There, \(\operatorname{MLP}\) stands for multi-layer perceptron and the absolute value is used to preserve the simmetry of the edge direction.

\subsection{Graph Distance or Similarity}

Following the idea of HED defined in Equation~\ref{eq:classicsoa:hed}, we propose to dynamically adapt the insertions and deletion costs according to the application domain. With this aim, we introduce two learnable costs \(c(\varepsilon \to v)\) and \(c(u \to \varepsilon)\) for the insertion and deletion of nodes. Thus, taking advantage of the computed node embeddings, our nodes are enriched with information aggregated from their local context and, therefore, its importance within the graph. Thus, we propose to take advantage of this in order to define two neural networks \(\varphi_i(\cdot)\) and \(\varphi_d(\cdot)\), defined as \(\varphi_* \colon \mathbb{R}^n \to \mathbb{R}^+\), able to decide the corresponding cost of this operation. In our experiments, \(\varphi_i(\cdot)\) and \(\varphi_d(\cdot)\) are the same network with shared weights as we consider the insertion and deletion operations to be symmetric. Moreover, we take the absolute value as the insertion and deletion costs must be positive.

Therefore, we define the distance between two graphs \(g_1=(V_1,E_1,\mu_1,\nu_1)\) and \(g_2=(V_2,E_2,\mu_2,\nu_2)\) as
\begin{equation}
    \begin{aligned}
    d_{\theta}(g_1,g_2) = \frac{1}{|V_1|+|V_2|}&\left( \sum_{u\in V_1 \cup \{\varepsilon\}} \min_{v\in V_2 \cup \{\varepsilon\}} c_\theta(u,v) \right. \\
    & \quad \left. + \sum_{v\in V_2 \cup \{\varepsilon\}} \min_{u \in V_1 \cup \{\varepsilon\}} c_\theta(u,v) \right),
    \end{aligned}
\end{equation}
where \(\theta\) are learnable parameters and \(c_\theta(\cdot,\cdot)\) is the corresponding learnable cost function defined as
\begin{equation}
    c_\theta(u,v) = 
    \begin{cases}
        \varphi_d(u;\theta) & \text{if } (u \to \varepsilon) \text{ is a deletion,} \\
        \varphi_i(v;\theta) & \text{if } (\varepsilon \to v) \text{ is an insertion,} \\
        \frac{d(u,v)}{2} & \text{otherwise.}
    \end{cases}
    \label{eq:learnedhed:costs}
\end{equation}

In our scenario, the edges are not taken into account as we consider the local structures to be already encoded during the message passing phase. However, edges can be incorporated to  Equation~\ref{eq:learnedhed:costs}, considering the adjacent edges as nodes and applying the same distance \(d_\theta(\cdot)\) with different learned weights.

Observe that an important aspect of the proposed distance is the fact that the node correspondence might not be symmetric. Figure~\ref{f:learningdistance:assignment}  illustrates this issue, moreover, we also show the effect of considering insertions and deletions as a \(\varepsilon\) node in Figure~\ref{f:learningdistance:assignment}(b). Note that not considering \(\varepsilon\) nodes as proposed in~\cite{riba2018learning} and illustrated in Figure~\ref{f:learningdistance:assignment}(a), can limit the learning capabilities of the framework.

\bgroup
\def\arraystretch{1}
\begin{figure}
     \centering
     \begin{tabular}{c@{\qquad}c}
        \includegraphics[height=0.19\textwidth, valign=c]{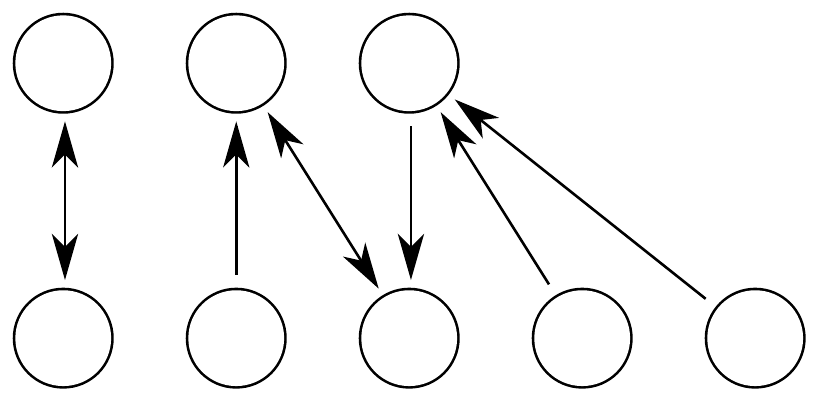} & \includegraphics[height=0.19\textwidth, valign=c]{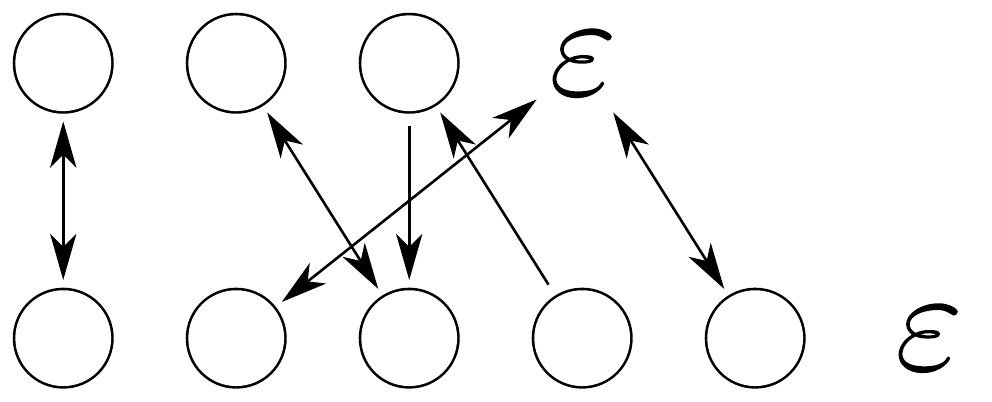} \\
        \textbf{(a)} & \textbf{(b)}
    \end{tabular}
    \caption{Assignment problem according to the proposed distance. (a) only substitutions are considered, (b) insertions and deletions are included as an extra epsilon node.}
    \label{f:learningdistance:assignment}
\end{figure}
\egroup

A limitation of our approach is that in some scenarios it might lose some node feature information in favor of encoding the local node structure. As a solution, we optionally combine the original graph information into Equation~\ref{eq:learnedhed:costs}, that can be redefined as,

\begin{equation}
    c_\theta(u,v) = 
    \begin{cases}
        \tau_d + \varphi_d(u;\theta) & \text{if } (u \to \varepsilon) \text{ is a deletion,} \\
        \tau_i + \varphi_i(v;\theta) & \text{if } (\varepsilon \to v) \text{ is an insertion,} \\
        \frac{d(u,v) + d'(u,v)}{2} & \text{otherwise,}
    \end{cases}
    \label{eq:learnedhed:costs_spatial}
\end{equation}
where \(\tau_d, \tau_i > 0\) are user-defined parameters to fix the minimum cost for node deletion and insertion respectively; \(d'(\cdot,\cdot)\) corresponds to the distance computed on the original node attributes. We find this setting helpful to avoid incorrect matchings between nodes that share structurally similar neighborhoods.

\section{Training setting and learning objective}
\label{s:learn}
In this work, we follow the idea of triplet networks to exploit the ranking properties of the desired metric. Thus, we use three GNN models with shared weights following the architecture illustrated in Figure~\ref{fig:learningdistance:pipeline}.

Our model is trained in a supervised manner, so we know which pairs of graphs belong to the same class. Compared to other approaches, we do not require node assignments nor a pre-computed similarity score. All models were trained using the \emph{Adam} optimizer~\cite{kingma2014adam} with weight decay \ie~\(L_2\) regularization. The learning rate of \(0.001\) is multiplied by \(0.95\) every 5 epochs to decrease its value, and we applied early stopping to finish our training process.

\begin{figure}[t]
    \centering
    \includegraphics[width=0.6\textwidth]{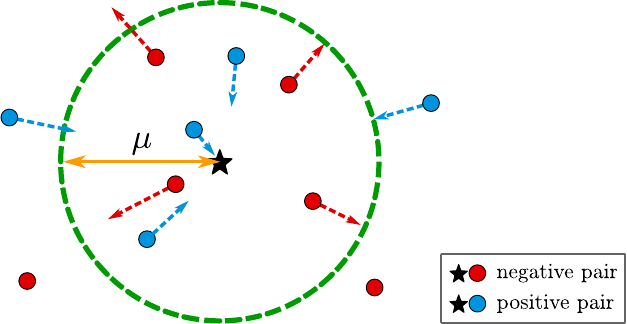}
    \caption{Illustration of the triplet learning objective. The anchor graph illustrated as a star should be close to its positive pair, in blue, and farther than a margin \(\mu\) to its negative counterpart.}
    \label{f:learningdistances:tripletobjective}
\end{figure}

The objective function to minimize is the triplet loss, also known as margin ranking loss. This learning objective receives three samples in which we already know its ranking, \ie~which pair should have a higher similarity score or distance. Let \(\{g_a, g_+, g_-\}\) be a triplet training sample where, \(g_a\) is the anchor graph, \(g_+\) is a positive graph sample \ie~a sample different from \(g_a\) but belonging to the same class and \(g_-\) is a negative graph example \ie~a sample belonging to a different class. Then, the triplet loss is defined as
\begin{equation}
    \mathcal{L}(\delta_{+}, \delta_{-}) = \max(0, \mu + \delta_{+} - \delta_{-}),
\end{equation}
where \(\mu\) is a fixed margin parameter, \(\delta_{+} = d_\theta(\phi(g_a), \phi(g_+))\) is the distance with respect to the positive sample and \(\delta_{-} = d_\theta(\phi(g_a), \phi(g_-))\)  is the distance with respect to the negative sample. Figure~\ref{f:learningdistances:tripletobjective} illustrates how this loss performs. Note that positive pairs are pushed to be close each other whereas negative samples are separated at least by the predefined margin \(\mu\).

Moreover, following the idea introduced in~\cite{balntas2016learning}, we apply an in-triplet hard negative mining which means that the anchor and positive samples can be swapped in case the positive sample is harder than the anchor one. Hence, we define \(\delta'_{-} = d_\theta(\phi(a), \phi(n))\) and \(\delta_{*} = \min(\delta_{-}, \delta'_{-})\). Finally, the new loss with the anchor swap is defined as

\begin{equation}
    \mathcal{L}(\delta_{+}, \delta_{*}) = \max(0, \mu + \delta_{+} - \delta_{*}).
    \label{eq:learningdistances:loss}
\end{equation}
Algorithm~\ref{alg:learningdistance:train} presents our training strategy. $\Gamma(\cdot)$ denotes the optimizer function.

\begin{algorithm}[t!]
    \caption{Training algorithm for our proposed model.} 
    \label{alg:learningdistance:train}
    \begin{algorithmic}[1]
        \REQUIRE Input data $\mathcal{G}$; max training iterations $T$  
        \ENSURE Networks parameters \(\Theta=\{\Theta_{\phi}, \theta\}\).
        \REPEAT
            \STATE Sample triplet mini-batches \(\{g_a, g_+, g_-\}_{i=1}^{N_B}\)
            \STATE $\mathcal{L} \leftarrow $ Eq.~\ref{eq:learningdistances:loss}
            \STATE $\Theta \leftarrow \Theta - \Gamma(\nabla_{\Theta}\mathcal{L})$ 
        \UNTIL{Convergence or max training iterations \(T\)}
    \end{algorithmic}
\end{algorithm}

\section{Experimental Validation}
\label{s:validation}

For validating our approach, a keyword spotting application in historical manuscripts has been considered as our main application scenario. Moreover, a final experiment on a classical mesh graph dataset is conducted. Our empirical evaluation demonstrate that the proposed approach provides competitive results when compared to the state-of-the-art. All the code is available at \url{github.com/priba/graph_metric.pytorch}.

\subsection{Historical Keyword Spotting}

In Document Image Analysis and recognition, Keyword Spotting (KWS), also known as word spotting, has emerged as an alternative to handwritten text recognition for documents in which the transcription performance is not satisfactory. Therefore, KWS is formulated as a content-based image retrieval strategy which relies upon obtaining a robust word image representation and a subsequent retrieval scheme.

\subsubsection{Dataset Description}

The HistoGraph dataset~\cite{stauffer2016novel,stauffer2018keyword} is a graph database for historical keyword spotting evaluation\footnote{Available at \url{http://www.histograph.ch/}}. It consists of different well known manuscripts.

\myparagraph{George Washington (GW)~\cite{fischer2012lexicon}:} This database is based on handwritten letters written in English by George Washington and his associates during the American Revolutionary War in 1755\footnote{George Washington Papers at the Library of Congress from 1741-1799, Series 2, Letterbook 1, pages 270-279 and 300-309, \url{https://www.loc.gov/collections/george-washington-papers/about-this-collection/}}. It consists of \(20\) pages with  a  total  of \(4,894\) handwritten words. Even tough several writers were involved, it presents small variations in style and only minor signs of degradation.

\myparagraph{Parzival (PAR)~\cite{fischer2012lexicon}:} This collection consists of \(45\) handwritten pages written by the German poet Wolfgang Von Eschenbach in the 13th century. The manuscript is written in Middle High German with a total of \(23,478\) handwritten words. Similarly to GW, the variations caused by the writing style are low, however, there are remarkable variations caused by degradation.

\myparagraph{Alvermann Konzilsprotokolle (AK)~\cite{pratikakis2016icfhr2016}:} It consists of German handwritten minutes of formal meetings held by the central administration of the University of Greifswald in the period of \(1794\) to \(1797\). In total \(18,000\) pages were used with small variations in style and only minor signs of degradation.

\myparagraph{Botany (BOT)~\cite{pratikakis2016icfhr2016}:} It consists of more than \(100\) different botanical records made by the government in British India during the period of \(1800\) to \(1850\). The records are written in English and contain certain signs of degradation and especially fading. The variations in the writing style are noticeable especially with respect to scaling and intra-word variations.

\bgroup
\def\arraystretch{2}
\setlength{\tabcolsep}{4pt}
\begin{figure}
     \centering
     \begin{tabular}{ccccc}
         \includegraphics[width=0.17\textwidth, valign=c]{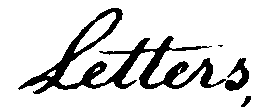} & \includegraphics[width=0.25\textwidth, valign=c]{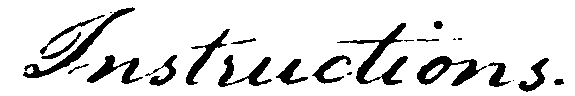} & \includegraphics[width=0.13\textwidth, valign=c]{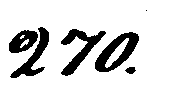} &  \includegraphics[width=0.15\textwidth, valign=c]{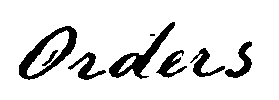} & \includegraphics[width=0.15\textwidth, valign=c]{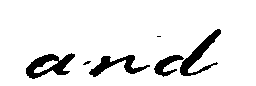} \\
        \multicolumn{5}{c}{\textbf{(a)} George Washington (GW)} \\
        
         \includegraphics[width=0.12\textwidth, valign=c]{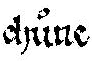} & \includegraphics[width=0.2\textwidth, valign=c]{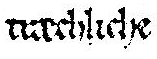} &
         \includegraphics[width=0.11\textwidth, valign=c]{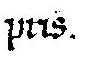} & \includegraphics[width=0.14\textwidth, valign=c]{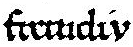} &  \includegraphics[width=0.12\textwidth, valign=c]{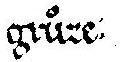} \\
        \multicolumn{5}{c}{\textbf{(b)} Parzival (PAR)} \\
        
        \includegraphics[width=0.17\textwidth, valign=c]{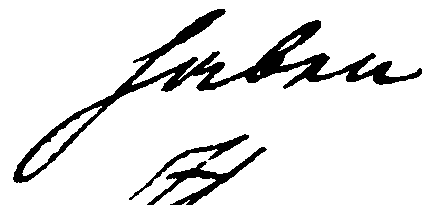} & \includegraphics[width=0.2\textwidth, valign=c]{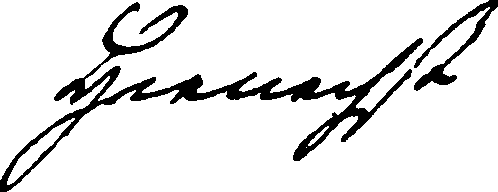} & \includegraphics[width=0.1\textwidth, valign=c]{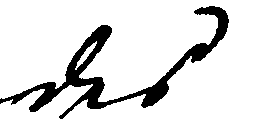} & \includegraphics[width=0.1\textwidth, valign=c]{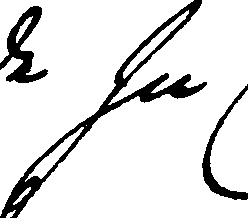} & \includegraphics[width=0.13\textwidth, valign=c]{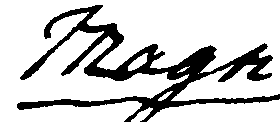} \\
        \multicolumn{5}{c}{\textbf{(c)} Alvermann Konzilsprotokolle (AK)} \\
        
        \includegraphics[width=0.2\textwidth, valign=c]{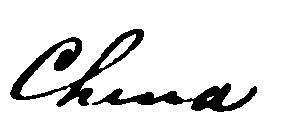} & \includegraphics[width=0.2\textwidth, valign=c]{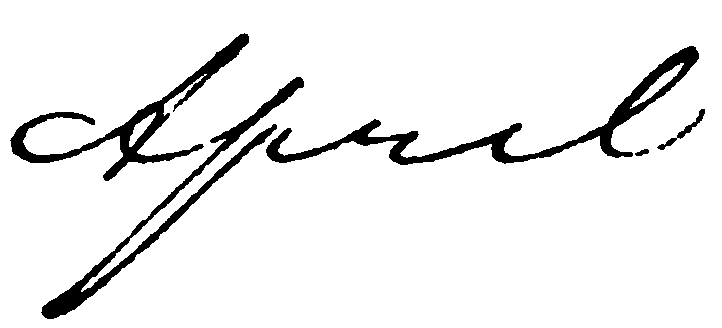} & \includegraphics[width=0.1\textwidth, valign=c]{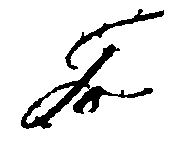} & \includegraphics[width=0.12\textwidth, valign=c]{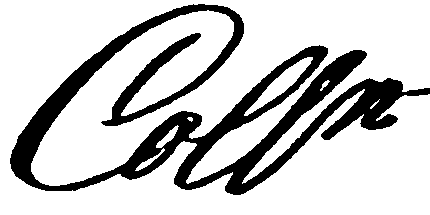} &  \includegraphics[width=0.12\textwidth, valign=c]{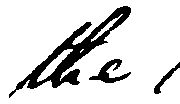} \\
        \multicolumn{5}{c}{\textbf{(d)} Botany (BOT)}
    \end{tabular}
    \caption{Pre-processed word examples of the four datasets.}
    \label{f:learningdistance:examples}
\end{figure}
\egroup

Figure~\ref{f:learningdistance:examples} provides some examples of pre-processed word images from which the graphs are created. Observe that the word segmentation of AK and BOT datasets is imperfect~\cite{pratikakis2016icfhr2016}. Moreover, these two datasets do not provide a validation set. So, some images from the training set have been used for validation. Table~\ref{t:datasetoverview} provides an overview of the dataset in terms of number of words.

\begin{table}[!htb]
    \centering
    \caption{Dataset overview in terms of number of keywords and word images for training, validating and testing respectively}
    \label{t:datasetoverview}
    \begin{tabular}{c c c c c}
        \toprule
        Dataset & Keywords & Train & Validation & Test \\
        \midrule
        GW & 105 & 2,447 & 1,224 & 1,224  \\
        PAR & 1,217 & 11,468 & 4,621 & 6,869 \\
        BOT & 150 & 1,684 & - & 3,380 \\
        AK & 200 & 1,849 & - & 3,734 \\
        \bottomrule
    \end{tabular}
\end{table}

To obtain a graph for each word in these datasets, the two most promising graph constructions introduced in~\cite{stauffer2018keyword} have been used:
\begin{itemize}
    \item \textbf{Keypoint}: Characteristic points are extracted from the skeletonized word image. Moreover between the connected characteristic points, equidistant nodes are inserted on top of the word skeleton.
    
    \item \textbf{Projection}: An adaptative grid is generated according to the vertical and horizontal projection profiles. Then, nodes are inserted in the corresponding center of mass of each grid cell. Moreover, undirected edges are inserted if nodes are directly connected by a stroke.
\end{itemize}

All the datasets presented in this work only contain spatial and structural information. This means that nodes are labelled with its normalized (x,y)-position on the image and that edges are unlabelled.

\subsubsection{Experimental protocol}

The experiments reported in this section use \(K=3\) GNN layers and, for the edit cost operation described in Equation~\ref{eq:learnedhed:costs_spatial}, we experimentally set the parameters \(\tau_i\) and \(\tau_d\) to \(0.5\).
Following the evaluation schemes of previous word spotting methodologies, we performed our experiments on two evaluation protocols. 

\begin{itemize}
    \item \textbf{Individual}: Each query image follows the traditional retrieval pipeline. Thus, queries are matched against the elements in the gallery and each ranking is evaluated independently.
    
    \item \textbf{Combined}: A query can consist of a set of graphs \(Q=\{q_1,\ldots,q_t\}\) where all the instances \(q\in Q\)  represent the same keyword. In this case, we consider the minimal distance achieved on all \(t\) query graphs. This second evaluation protocol was adopted in some previous graph-based word spotting works~\cite{ameri2019graph,stauffer2018keyword}. The motivation of this setting is to mitigate the structural bias provided by the query instance, \ie~different handwriting styles can provide extremely different graphs, so, in terms of graph distances, it is unrealistic to consider them from the same class. 
\end{itemize}

For the evaluation, we use the mean Average Precision (mAP), a classic information retrieval metric~\cite{rusinol2009performance}. First, let us define Average Precision (AP) as
\begin{equation}
    \operatorname{AP} = \frac{\sum_{n=1}^{|\operatorname{ret}|} P@n \times r(n)}{|\operatorname{rel}|},
\end{equation}
where \(P@n\) is the precision at \(n\) and \(r(n)\) is a binary function on the relevance of the \textit{n}-th item in the returned ranked list. Then, the mAP is defined as:
\begin{equation}
    \operatorname{mAP} = \frac{\sum_{q=1}^{Q} \operatorname{AP}(q)}{Q},
\end{equation}
where $Q$ is the number of queries.

\subsubsection{Ablation study}

We first empirically investigate the influence of the margin parameter \(\mu\) to each model, as well as the importance of the GNN layers choice. Table~\ref{t:histographablation} presents a comparison of the different settings, providing the averaged mAP of 4 runs and its corresponding standard deviation. This evaluation has been done for all the datasets and for both graph representations~\viz~Keypoint and Projection. The evaluation protocol in this experiment is Individual as we believe that it is the natural experimental setting for this problems.

From these results, we observe that GRU-based models are slightly better, allowing a higher degree of deformations between words from the same class. Note that both datasets AK and BOT do contain imperfect word segmentations. In addition, these datasets contain samples written with a more artistic calligraphic style as shown in Figure~\ref{f:learningdistance:examples}. Thus, the artistic strokes are drivers of a higher degree of complexity. Observe that the performance drop in BOT dataset can be also explained by the artistic nature of the dataset.

Additionally, the Keypoint representation performs the best on the GW dataset, since the strokes are simpler and its structure in terms of the binary image skeleton is more relevant for a proper retrieval. Finally, we observe that a higher margin \(\mu\) is more adequate for the GAT-based models whereas it's harmful for the GRU-based one.

\bgroup
\setlength{\tabcolsep}{3.25pt}
\begin{table}[!htb]
    \centering
    \caption{Study on the GNN model and margin parameter of the proposed model. Mean average precision (mAP) and standard deviation (average on four runs) for graph-based KWS system on George Washington (GW), Parzival (PAR) Alvermann Konzilsprotokolle (AK) and Botany (BOT) datasets.}
    \begin{tabular}{l l l c c c c c c c c c c c c}
        \toprule
         & \multirow{2}{*}{Model} & \multirow{2}{*}{\(\mu\)} && \multicolumn{2}{c}{GW} && \multicolumn{2}{c}{PAR} && \multicolumn{2}{c}{AK} && \multicolumn{2}{c}{BOT} \\
        \cmidrule{5-6} \cmidrule{8-9} \cmidrule{11-12} \cmidrule{14-15}
        &  & && mAP & \(\pm\) && mAP & \(\pm\) && mAP & \(\pm\) && mAP & \(\pm\) \\
        \midrule
        \multirow{4}{*}{\rotatebox{90}{\textbf{Keypoint}}} & GAT & 1 && 72.49 & 1.169 && 66.46 & 3.162 && 62.90 & 1.325 && 39.86 & 0.396\\
            &  & 10  && \textbf{76.92} & 2.309 && \textbf{73.14} & 0.973 && 62.72 & 1.783 && \textbf{41.52} & 0.782 \\
        \cmidrule{2-15}
        & GRU & 1 && 72.86 & 3.331 && 67.27 & 1.281 && \textbf{64.42} & 1.003 && 39.69 & 0.532 \\
            &  & 10 && 68.45 & 2.715 && 48.59 & 9.571 && 60.84 & 1.127 && 38.22 & 0.778 \\
        \midrule
        \multirow{4}{*}{\rotatebox{90}{\textbf{Projection}}} & GAT & 1 && 67.86 & 2.379 && 70.77 & 1.906 && 63.44 & 1.233 && 39.12 & 2.037 \\
            &  & 10 && \textbf{70.25} & 3.431 && \textbf{75.19} & 0.755 && 62.72 & 1.518 && 38.83 & 2.801 \\
        \cmidrule{2-15}
        & GRU & 1 && 68.09 & 1.234 && 71.07 & 1.933 && \textbf{65.04} & 1.226 && \textbf{42.83} & 0.568 \\
            &  & 10 && 63.39 & 4.222 && 52.32 & 1.298 && 60.51 & 1.451 && 37.59 & 0.778 \\
        \bottomrule
    \end{tabular}
    \label{t:histographablation}
\end{table}
\egroup

\subsubsection{Results and discussion}

Table~\ref{t:histographcomparison} compares with graph-based methodologies. In this setting we follow the Combined evaluation protocol reported by~\cite{ameri2019graph,stauffer2018keyword}. For each dataset and graph representation, we use the best model reported in the previous section. Observe that, for a fair comparison, that is, using the same graph representation, we outperform both AED~\cite{riesen2009approximate} and HED~\cite{fischer2015hausdorff} on all the datasets but AK, where we obtain very similar results. However, the ensemble methods reported in~\cite{stauffer2018keyword} are able to obtain a better performance on the datasets with more variability. Note that these ensembles combine, in different ways, the graph distances computed on different graph representations of the same images. Therefore, we do not consider it a fair comparison but a remarkable fact that the performance of our system is able to outperform them in two of the datasets while obtaining competitive results on the other two.

\begin{table}[!htb]
    \centering
    \caption{State-of-the-art on graph-based KWS techniques. Mean average precision (mAP) for graph-based KWS system on GW, PAR, AK and BOT datasets.}
    \begin{tabular}{l ll@{\hskip 48pt} c@{\hskip 24pt} c@{\hskip 24pt} c@{\hskip 24pt} c}
        \toprule
        Distance & \multicolumn{2}{l}{Representation} & GW & PAR & AK & BOT \\
        \midrule
        AED~\cite{riesen2009approximate} & \multicolumn{2}{l}{Keypoint~\cite{ameri2019graph}}   & 68.42 &55.03 & 77.24 & 50.94 \\
            & \multicolumn{2}{l}{Projection~\cite{ameri2019graph}} & 60.83 & 63.35 & 76.02 & 50.49 \\
            & \multirow{5}{*}{\rotatebox{90}{Ensemble~\cite{stauffer2018keyword}}} &\(\min\) & 70.56 & 67.90 & 82.75 & 65.19  \\
            & & \(\max\) & 62.58 & 67.57 & 82.09 & 67.57 \\
            & & \(\operatorname{mean}\) & 69.16 & 79.38 & 84.25 & \textbf{68.88} \\
            & & \(\operatorname{sum}_{\alpha}\) & 68.44 & 74.51 & \textbf{84.77} & 68.77 \\
            & & \(\operatorname{sum}_{\operatorname{map}}\) & 70.20 & 76.80 & 84.25 & \textbf{68.88} \\
        \midrule
        HED~\cite{fischer2015hausdorff} & \multicolumn{2}{l}{Keypoint~\cite{ameri2019graph}}   & 69.28 & 69.23 & 79.72 & 51.74 \\
            & \multicolumn{2}{l}{Projection~\cite{ameri2019graph}} & 66.71 & 72.82 & \textbf{81.06} & 51.69 \\
        \midrule
        \textbf{Ours} & \multicolumn{2}{l}{Keypoint}   & \textbf{78.48} & 79.29 & 78.64 & 51.90 \\
            & \multicolumn{2}{l}{Projection} & 73.03 & \textbf{79.95} & 79.55 & \textbf{52.83} \\
        \bottomrule
    \end{tabular}
    \label{t:histographcomparison}
\end{table}

Table~\ref{t:histographAKBOT:statistical} shows a comparison with non-graph based approaches. In particular, we compare against three state-of-the-art learning-based reference systems of the ICFHR2016 competition~\cite{pratikakis2016icfhr2016}. In this case, the evaluation of these learning-based frameworks follows the protocol described in the competition. Thus, queries of the same query keyword are considered to be independent. Note that, due to this query protocol, the learning-based frameworks are not directly comparable to the state-of-the-art graph-based KWS results that we have reported above. In this table, we can observe the superiority of learning methods working directly on the image domain. In particular, PHOCNet~\cite{sudholt2016phocnet} leads to stunning accuracies for this task. However, the proposed graph-based approach, is able to provide a new step towards closing the gap between structural and statistical methodologies on this kind of tasks. In addition, the proposed approach is able to deal with the noise introduced by the graph construction.

\begin{table}[!htb]
    \centering
    \caption{Comparison against non-graph learning based systems. Mean average precision (mAP) for graph-based KWS system on AK and BOT datasets.}
    \begin{tabular}{l l@{\hskip 24pt} c@{\hskip 24pt} c}
        \toprule
       Method & Representation & AK & BOT \\
        \midrule
        CVCDAG~\cite{almazan2014PAMI} & - & 77.91 & 75.77\\
        PHOCNet~\cite{sudholt2016phocnet} & - & 96.05 & 89.69\\
        QTOB~\cite{wilkinson2016semantic} & - & 82.15 & 54.95\\
        \midrule
        \textbf{Ours} & Keypoint   & 64.42 & 41.52\\
            & Projection & 65.04 & 42.83\\
        \bottomrule
    \end{tabular}
    \label{t:histographAKBOT:statistical}
\end{table}

Finally, Figure~\ref{f:learningdistance:qualitative} provides qualitative examples of our matching framework for a positive and negative sample. The first row provides the top to bottom matching whereas the second row shows the opposite, from the bottom to the top. Notice that in the positive sample case, both directions are much more consistent than in the negative sample case. 

\bgroup
\def\arraystretch{1.5}
\begin{figure}
     \centering
     \begin{tabular}{c@{\hskip 0.5em} c}
        \includegraphics[width=0.474\textwidth, valign=c]{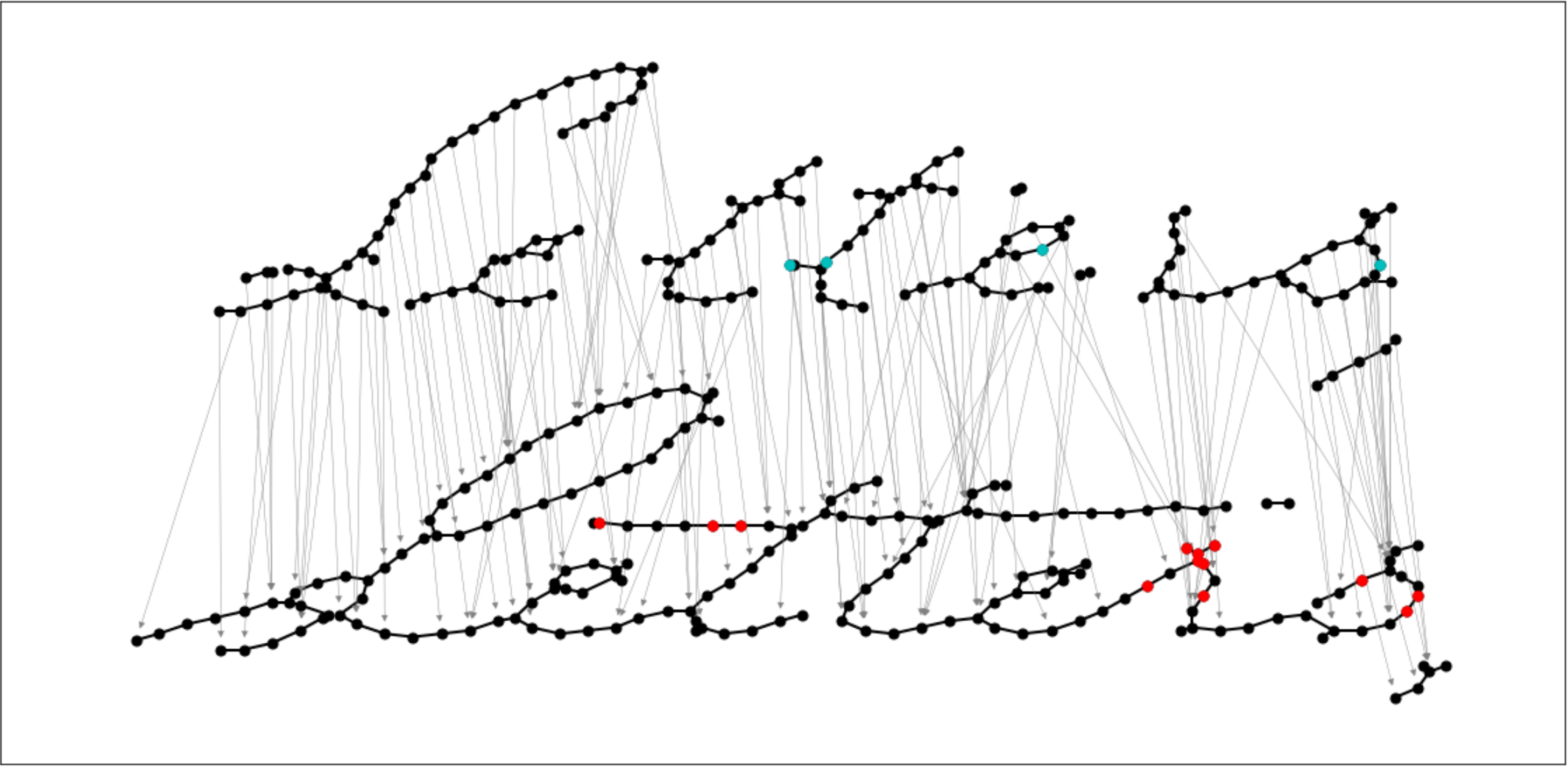} & \includegraphics[width=0.474\textwidth, valign=c]{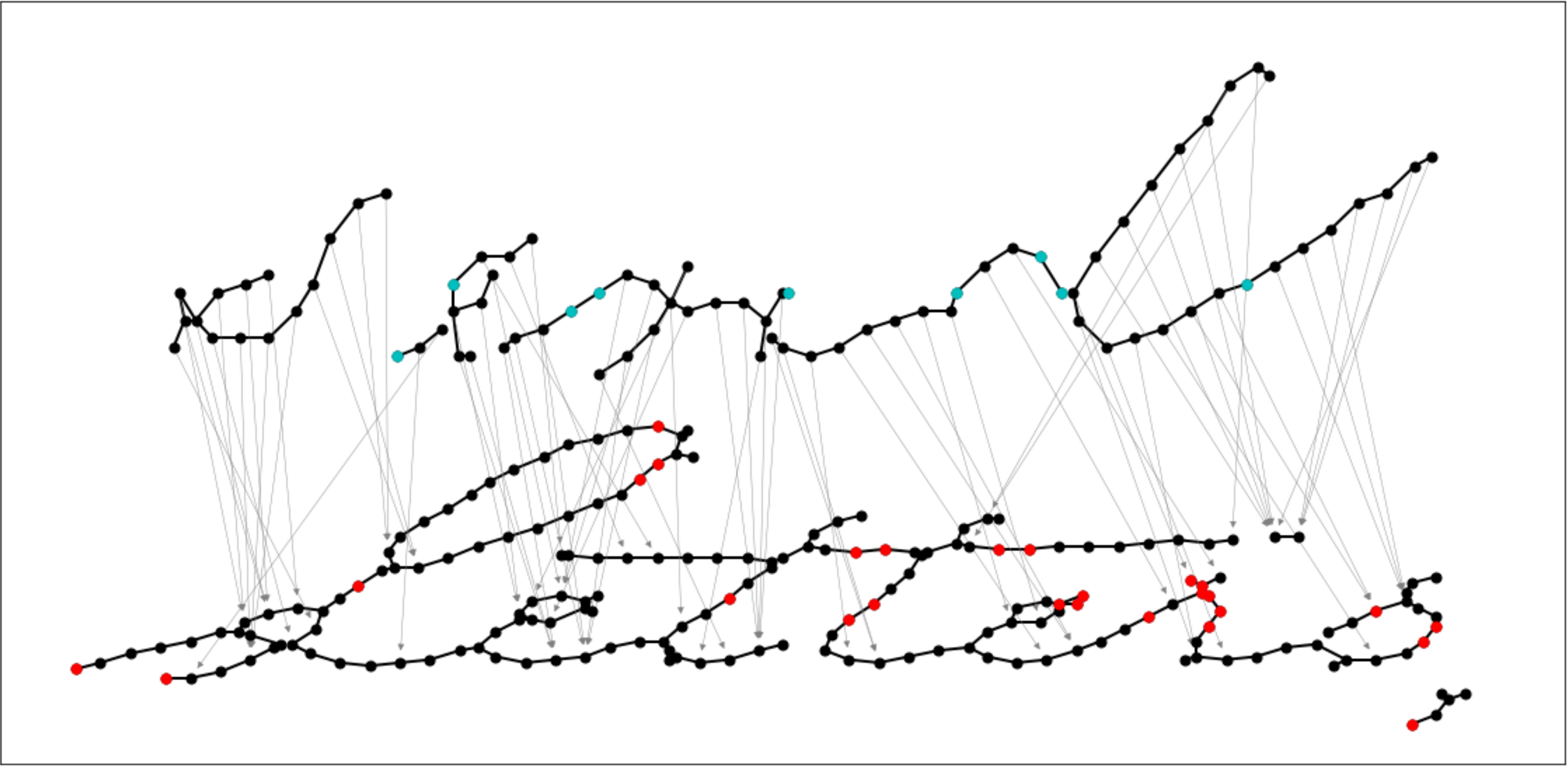} \\
        \textbf{(a)} & \textbf{(b)} \\
        \includegraphics[width=0.474\textwidth, valign=c]{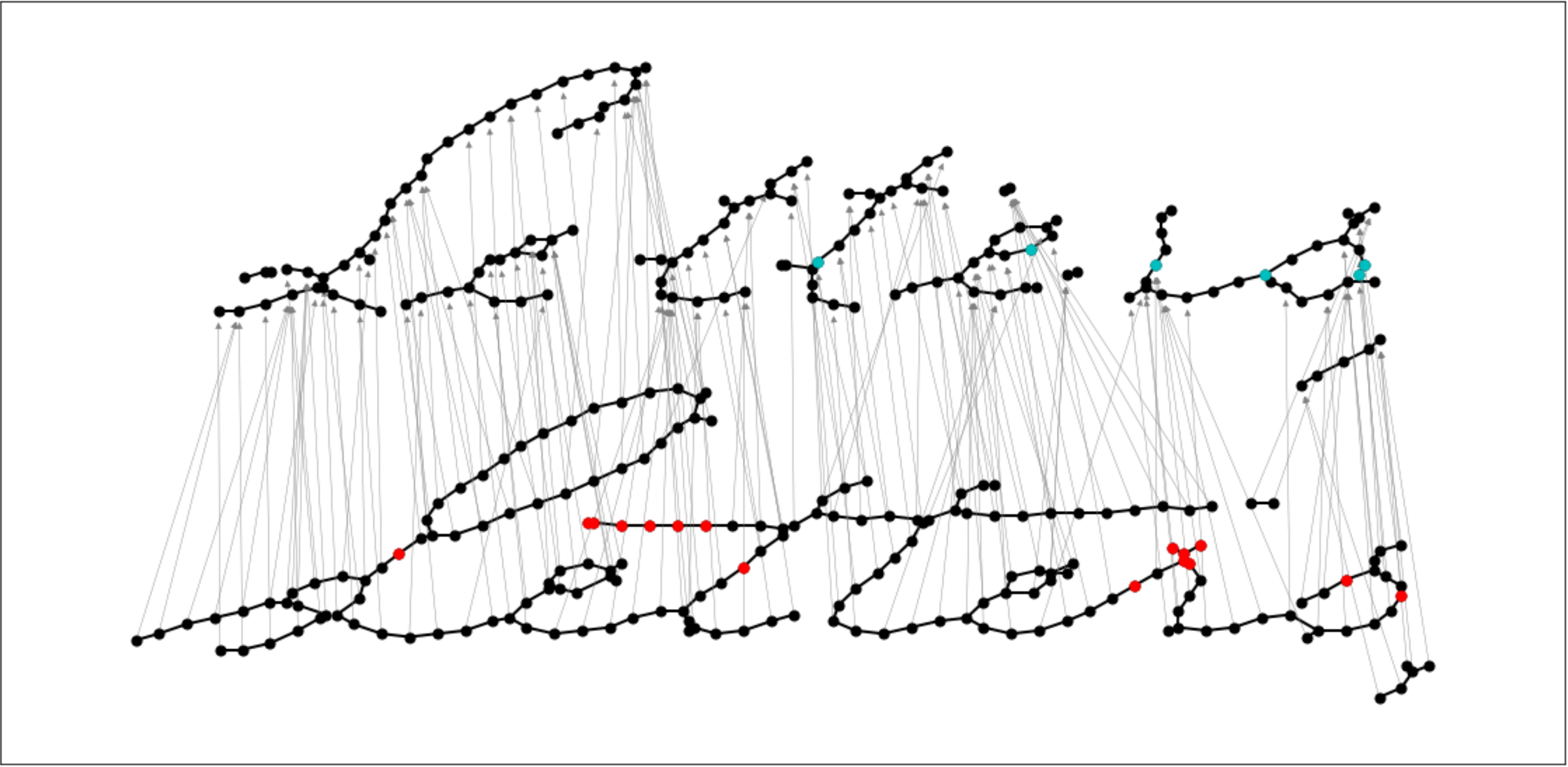} & \includegraphics[width=0.474\textwidth, valign=c]{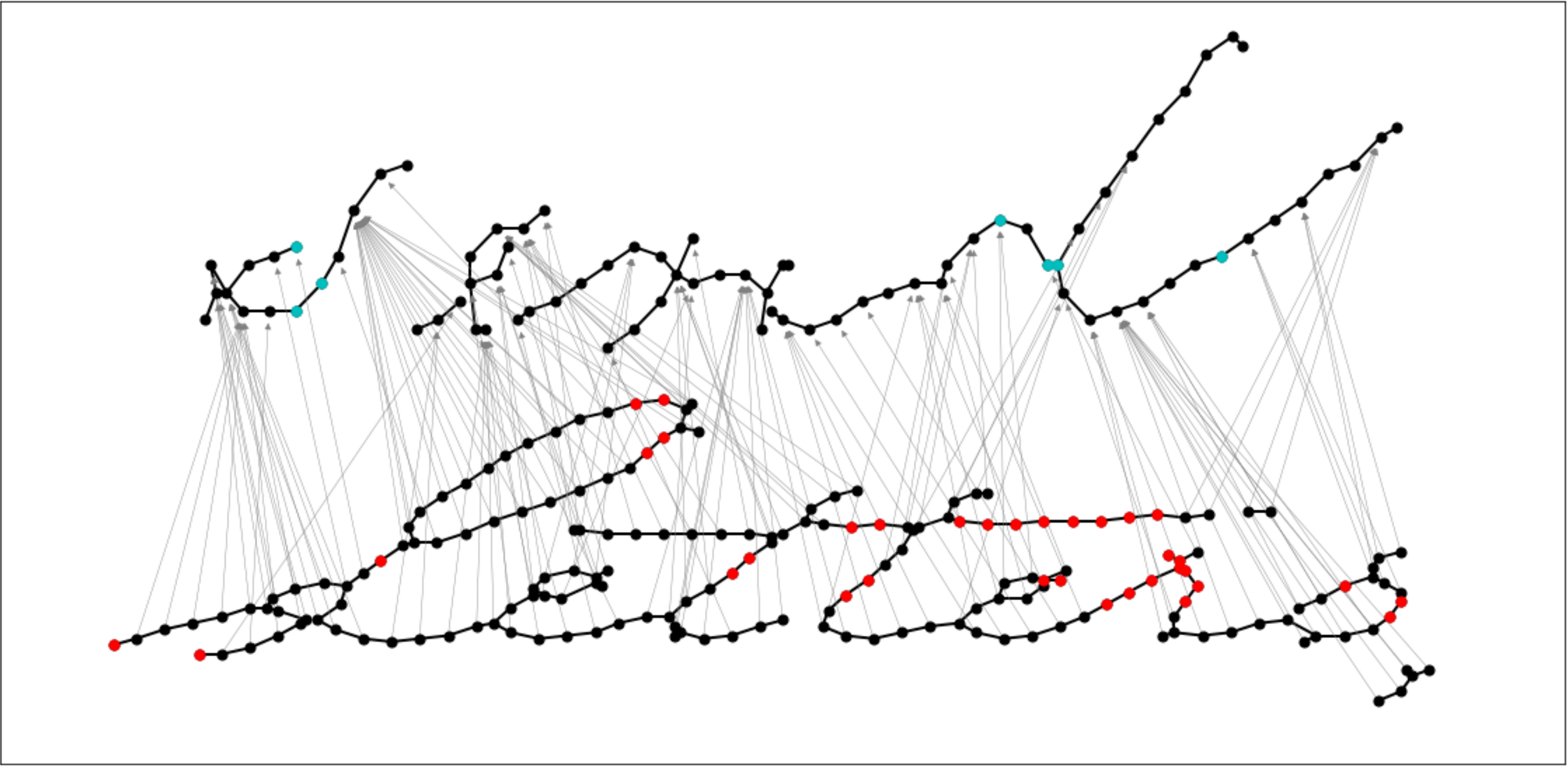} \\
        \textbf{(c)} & \textbf{(d)} 
    \end{tabular}
    \caption{Visualization of the learned node correspondence. First row, shows the node matching from top to bottom; the second row of the figure shows de opposite. (a)-(d) ``Letters''--``Letters''; (b)-(d) ``send''--``Letters''}
    \label{f:learningdistance:qualitative}
\end{figure}
\egroup

\subsection{Experimental comparison to GMN}

Among the graph metric learning approaches in the literature, the graph matching networks (GMN) work~\cite{li2019graph} is the most prominent one. In this section, we propose an extra experiment to compare with their work.

\subsubsection{Dataset Description}

The IAM Graph Database Repository~\cite{Riesen2008} provides several graph datasets covering a wide spectrum of different applications. In particular, we focused on the COIL-DEL dataset. The COIL-100~\cite{nayar1996columbia} consists of 100 object images at different poses. In order to construct the COIL-DEL dataset, these images were converted into mesh graphs by means of the Harris corner detection algorithm followed by a Delaunay triangulation. COIL-DEL is divided in \(2,400\), \(500\) and \(1,000\) graphs for training, validation and test respectively. In average these graphs have \(21.5\) nodes and \(54.2\) edges. Thus, they are rather small graphs.

\subsubsection{Experimental protocol}

In these experiments, we followed the same experimental protocol introduced by Li~\etal~\cite{li2019graph}, so we evaluated our method on two different metrics:

\begin{itemize}
    \item \textbf{pair AUC}: The area under the ROC curve for classifying pairs of graphs as similar or not on a fixed set of 1,000 pairs.
    \item \textbf{triplet accuracy}: The accuracy of correctly assigning a higher similarity to the positive pair than a negative pair on a fixed set of 1,000 triplets.
\end{itemize}

Note that the fixed pairs and triplets are not the same from the original paper. We performed a random selection of these pairs and triplets while trying to balance the number of examples per class.

\subsubsection{Results and Discussion}

Table~\ref{t:learningdistances:gmncomparison} presents a comparison to the state-of-the-art techniques on graph metric learning. In particular, we compare with the different architectures proposed in~\cite{li2019graph}. 

It is not surprising that their GMN technique outperforms our proposed model as they do incorporate cross-graph connections following an attention paradigm. Therefore, the correspondence is learned end-to-end in a much robust way. However, this is only feasible in rather small datasets as it incorporates a huge computational overhead. Notice also, that their GNN and Siamese-GNN models just obtain a slightly better performance than our proposed approach. However, when dealing with such small graphs it is hard to compare against embedding based approaches as they are able to encode the graph characteristic features without a huge loss of information.

In this extra experiment, we have also evaluated the effect on the choice of GNN models, the number of layers and the margin parameter \(\mu\). From this table, we conclude that the GRU-based models are drivers of a better performance on these experiments. Moreover, we find important to set our margin parameter to \(1\). The number of layers has not proven to bring a boost on performance on this particular dataset. Overall, our best model is able to obtain comparable results against GMN in this small dataset.

\begin{table}[htb]
    \centering
    \caption{Performance comparison on the COIL-DEL dataset against the methodologies introduced in~\cite{li2019graph}. We studied the effect of the proposed GAT and GRU models, as well as, the number of layers and margin parameter \(\mu\).}
    \begin{tabular}{l l l l c c}
    \toprule
        \multicolumn{2}{l}{Model} & \(\#\) Layers (\(K\)) & \(\mu\) & Pair AUC & Triplet Acc \\
        \midrule
        \multicolumn{2}{l}{GCN~\cite{li2019graph}} & - & - & 94.80 & 94.95 \\
        \multicolumn{2}{l}{Siamese-GCN~\cite{li2019graph}} & - & - & 95.90 & 96.10 \\
        \multicolumn{2}{l}{GNN~\cite{li2019graph}} & - & - & 98.58 & 98.70 \\
        \multicolumn{2}{l}{Siamese-GNN~\cite{li2019graph}} & - & - & 98.76 & 98.55 \\
        \multicolumn{2}{l}{GMN~\cite{li2019graph}} & - & - & 98.97 & 98.80 \\
    \midrule
        \textbf{Our} & GAT & 3 & 1 & 97.82 & 96.74 \\
           & &   & 10 & 96.22 & 96.94 \\
        \cmidrule{3-6}
            & & 5 & 1 & 97.85 & 96.94 \\
            & &   & 10 & 97.70 & 97.54 \\
        \cmidrule{2-6}
         & GRU  & 3 & 1   & 98.08 & 97.50 \\
            &&   & 10  & 96.25 & 95.69 \\
        \cmidrule{3-6}
            && 5 & 1 & 97.56 & 97.60 \\
            &&   & 10 & 95.36 & 96.07 \\
    \bottomrule
    \end{tabular}
    \label{t:learningdistances:gmncomparison}
\end{table}

\section{Conclusions and Future Work}
\label{s:conc}

In this paper, we have proposed a triplet GNN architecture for learning graph distances. Our architecture is able to learn node embeddings based on structural information of nodes local contexts. These learned features lead to an enriched graph representation which is later used in the distance computation. Moreover, we extended the graph edit distance approximation \viz~Hausdorff edit distance, to the new learning framework in order to learn its operation costs within an end-to-end fashion. We have validated our proposed architecture on a graph retrieval scenario, in particular, we faced a keyword spotting task for handwritten words. Finally, we demonstrated competitive results against state-of-the-art, learning-based methods for graph distance learning.

Several future research lines emerge taking our proposed framework as starting point. For instance, our framework does not exploit the edge structure at matching time as we considered it implicitly encoded as node features. However, leveraging the edges information at this stage might lead to better results. Another promising line of research relates to the use of different graph pooling layers for reducing the size of large graphs before computing the learned Hausdorff edit distance.


\section*{Acknowledgment}

This work has been partially supported by the Spanish project RTI2018-095645-B-C21, the FPU fellowship FPU15 / 06264, the Ramon y Cajal Fellowship RYC-2014-16831, and the CERCA Program / Generalitat de Catalunya. We thank NVIDIA for the donation of a Titan Xp GPU.

\section*{References}

\bibliography{bib/thesis.bib,bib/pub.bib}

\end{document}